\def\year{2022}\relax
\documentclass[letterpaper]{article} 
\usepackage{aaai22}  
\usepackage{times}  
\usepackage{helvet}  
\usepackage{courier}  
\usepackage[hyphens]{url}  
\usepackage{graphicx} 
\usepackage{natbib}  
\usepackage{caption} 
\DeclareCaptionStyle{ruled}{labelfont=normalfont,labelsep=colon,strut=off} 
\usepackage{subcaption}
\usepackage{algorithm}
\usepackage{algorithmic}
\usepackage{amssymb}
\usepackage{amsmath}
\usepackage{amsfonts}
\usepackage{booktabs}
\usepackage{threeparttable}
\usepackage{multirow}
\usepackage{rotating}

\title{Deep Movement Primitives: toward Breast Cancer Examination Robot}
\author{
    Oluwatoyin Sanni,\textsuperscript{\rm 1,*}
    Giorgio Bonvicini,\textsuperscript{\rm 2}
    Muhammad Arshad Khan,\textsuperscript{\rm 1}
    Pablo C. L\'opez-Custodio,\textsuperscript{\rm 1}
    Kiyanoush Nazari,\textsuperscript{\rm 1}
    Amir M. Ghalamzan E.\textsuperscript{\rm 1,}\footnote{Authors contributed equally.} 
}
\affiliations {
    \textsuperscript{\rm 1} University of Lincoln, UK, 
    \textsuperscript{\rm 2} Polytechnic University of Milan, Italy\\
    aghalamzanesfahani@lincon.ac.uk
}
 
\begin{document}

\maketitle
\begin{abstract}
Breast cancer is the most common type of cancer worldwide. A robotic system performing autonomous breast palpation can make a significant impact on the related health sector worldwide. However, robot programming for breast palpating with different geometries is very complex and unsolved. Robot learning from demonstrations (LfD) re- duces the programming time and cost. However, the available LfD are lacking the modelling of the manipulation path/trajectory as an explicit function of the visual sensory information. This paper presents a novel approach to manipulation path/trajectory planning called deep Movement Primitives that successfully generates the movements of a manipulator to reach a breast phantom and perform the palpation. We show the effectiveness of our approach by a series of real-robot experiments of reaching and palpating a breast phantom. The experimental results indicate our approach outperforms the state-of-the-art method.  
\end{abstract}

Breast cancer is the most common cancer worldwide with a significant impact on the life of patients and society~\cite{cancerwho}. 2.3 million females were diagnosed with breast cancer in 2020 with 685,000 related deaths worldwide \cite{sung2021global}. Less invasive and cheaper cancer treatments and higher quality of patients' lives post-cancer detection are among the benefits of early breast cancer detection~\cite{kosters2003regular}. Breast Palpation (\textbf{BP}) is the easiest, most effective and most widely used early cancer detection method. BP -- both self and clinical examinations -- seeks to detect palpable anomalies in the breast tissue~\cite{provencher2016clinical, saslow2004clinical}. BP involves human tactile and visual inspection during palpating the breast and lymph nodes. Because of the lack of patients' expertise in palpation, self-examination is ineffective across societies. Moreover, subjects are reluctant to be examined by human experts~\cite{yang2010motivations}, detection precision depends on the examiner's expertise, and the availability of experts everywhere may be limited. Therefore, our survey shows autonomous robotic BP (\textbf{ARBP}) is of great interest to both clinicians and patients~\cite{artemissurvey2020}

ARBP reduces the burden on the general healthcare system for clinical BP, improves the accessibility of the service and precision of early breast cancer detection. Although there are many studies for robotic tissue palpation for tumour detection \cite{scimeca2020structuring, herzig2018variable, kobayashi2009robotic, nichols2015methods, keshavarz2015dynamic}, most of them focus on tissue stiffness classification and developing a suitable robot finger/hand for palpation. Hence, the problem of efficient path/motion planning for autonomous robotic palpation remains vastly unexplored. Geometrical variability across different palpation paths and the subjects' breasts leads to the complexity of motion/path planning for breast palpation. 

\begin{figure}[tb!]
    \centering
    \begin{subfigure}{0.3\textwidth}    
        \centering
        \includegraphics[width=1\columnwidth, scale=0.9]{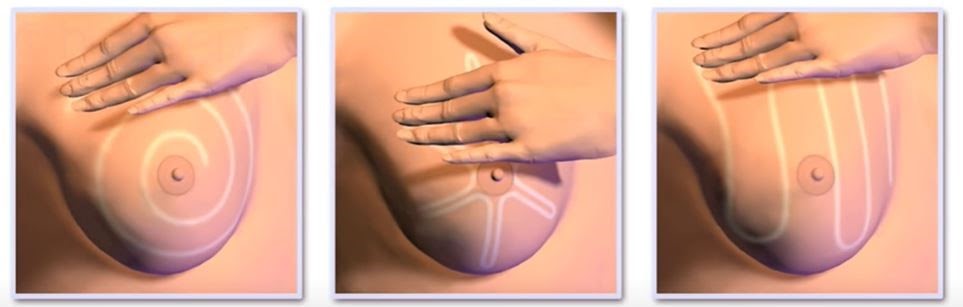}
        \caption{\small Palpation patterns~\cite{palpatterns}}
        \label{fig:patterns}
    \end{subfigure}
    \\
    \begin{subfigure}{1\columnwidth}
        \centering
        \includegraphics[width=1\linewidth]{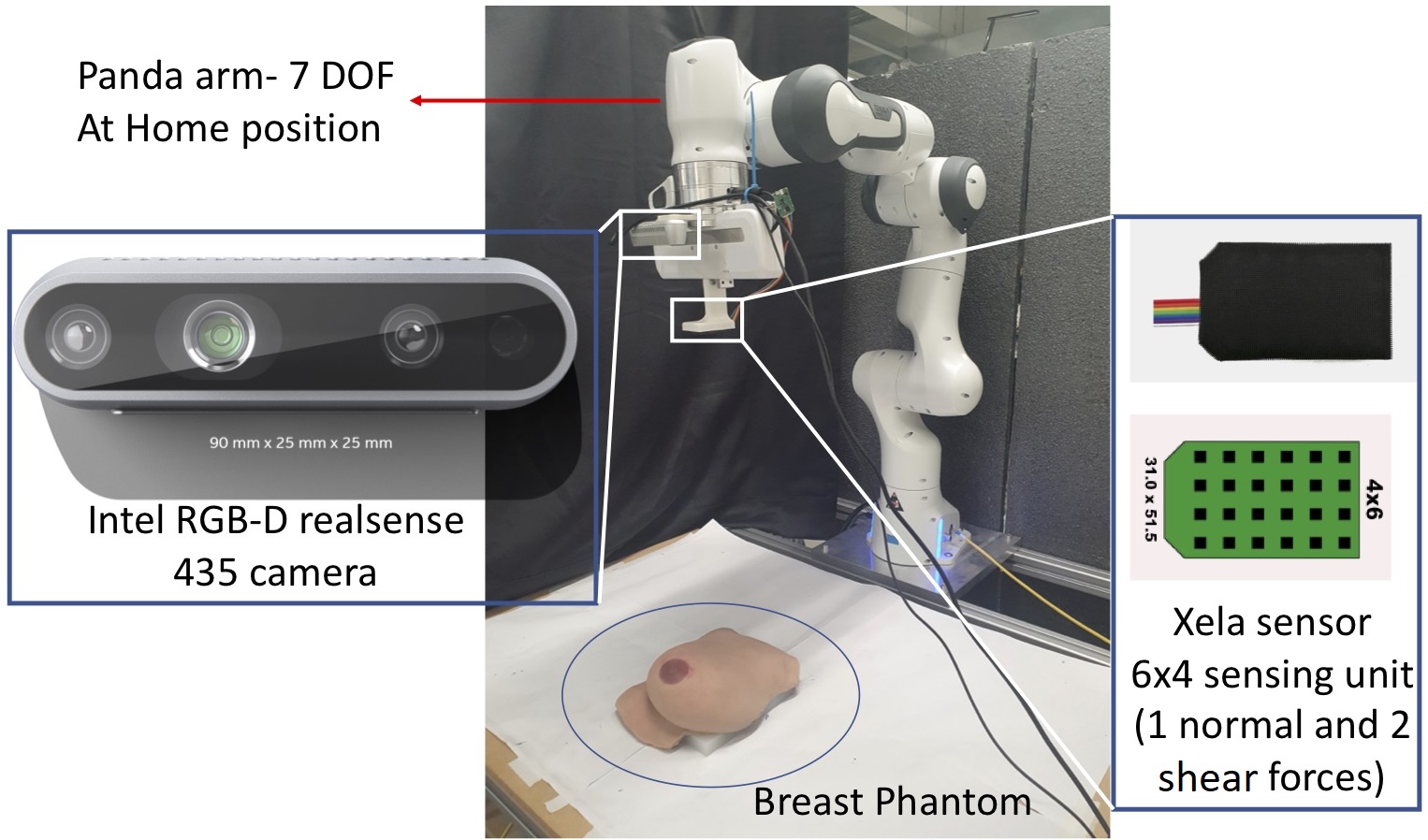}
        \caption{{\small Panda arm at home configuration}}
        \label{fig:setup0}
    \end{subfigure}%
    \\
    \begin{subfigure}[b]{0.2\textwidth}
        \centering
        \includegraphics[height=3.7cm]{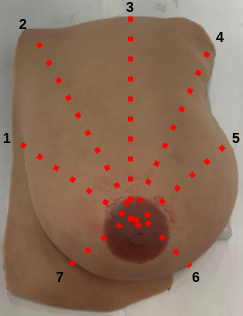}
        \caption{Breast phantom}
        \label{fig:wedgepattern}
    \end{subfigure}
    \begin{subfigure}[b]{0.2\textwidth}
        \centering
        \includegraphics[height=3.7cm]{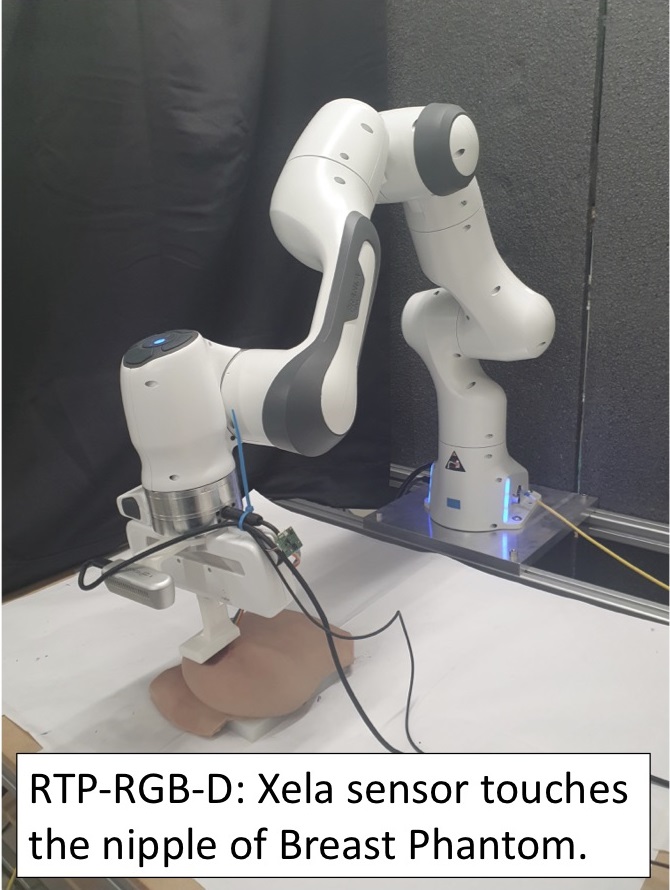}
        \caption{Reach-to-palpate}
        \label{fig:rtp_rgbd}
    \end{subfigure}
    \caption{ \small{Breast palpation is an efficient and the first approach to evaluate glandular tissue and nipple~\cite{mahoney1982efficiency}: (a) well-known breast palpation paths: Spiral (left), Wedges (middle) and Lines (right); (b) 7-DOF Panda arm at Home Configuration; (c) the breast phantom used use for study in this paper and the demonstrated palpation paths. The phantom is made of silicon with 2 lumps implanted on path 4 and 7}; (d) reach-to-palpate (RTP) action of Panda arm; RGB-D images taken at home configuration shown in Fig.~\ref{fig:setup0}.}
    \label{fig:palp}
\end{figure}

Human experts suggest benchmark patterns for a more effective way of performing breast palpation (see Fig.~\ref{fig:patterns}), such as circular, wedges or linear~\cite{murali1992comparison}. An autonomous robotic palpation system needs to encode the expert knowledge into the palpation movements and adapt the planned motions according to the breast geometry. 
Movement Primitives as a tool for compact representation of robot's control policy are used for generating robot motions~\cite{ude2010task, matsubara2011learning, ragaglia2018robot}. Probabilistic Movement Primitives (\textbf{ProMP}) \cite{paraschos2013probabilistic} can be used for planning and control purposes where it can express a distribution of trajectories by the corresponding variance and mean. Nonetheless, these conventional Movement Primitives methods~\cite{paraschos2018using,paraschos2013probabilistic, schaal2006dynamic,ude2010task} cannot capture the correlation between the visual information and the generated movements.

We propose a novel Learning from Demonstration (\textbf{LfD}) approach called \textit{deep movement primitives} (\textbf{deep-MP})\footnote{Code, data and an extended appendix are available here:\\\url{https://github.com/imanlab/deep_movement_primitives}} directly mapping the visual sensory information into the learned trajectory. 
Other contributions of this paper include: (i) the effectiveness of deep-MP variations for different tasks complexities are extensively studied; (ii) a series of real-robot breast palpation experiments that show the effectiveness of our proposed approach to complex trajectory/path planning tasks.

\section{Related Works}\label{sec:relworks}

In trajectory planning, LfD methods use demonstrations to build a task model and execute it.
Probabilistic approaches such as Gaussian Mixture models (GMM)~\cite{jaquier2019learning, girgin2020active}, Gaussian Process~\cite{schneider2010robot} are applied in LfD settings for expressing the distributions of trajectories by the corresponding mean and covariance. However, these models do not encode the relation between the visual information of robot's workspace and the demonstrated behaviour~\cite{rana2018towards}. 

End-to-end learning-based control approaches such as Inverse Reinforcement Learning \cite{levine2011nonlinear}, vision-based Model Predictive Control \cite{finn2017deep}, or Behaviour cloning \cite{rahmatizadeh2018vision} usually suffer from lack of generalisation and they are only applied to a class of tasks involved complicated motion control but not complex motion planning.
End-to-end LfD enables generating control commands directly from raw image data, e.g., 
~\cite{rahmatizadeh2018vision} proposed behaviour cloning probabilistic generative model. 
However, high dimensional observation space, e.g. raw images, make the combined motion/path planning and motion control intractable in such settings~\cite{akgun2012keyframe, nagahama2019learning}. In a line of research, deep-time-series are used to learn the control policy of a robot to perform a specific task. For instance, \cite{levine2018learning, levine2016end} applied Deep Neural Networks (NNs) combined with Recurrent NNs to image data to learn robots' control policy. In these frameworks, deep NNs finds a mapping between the raw image and desired control signals. These approaches are proved to be effective only for limited classes of tasks with challenging motion control. Moreover, such models are hard to interpret.

Dynamic Movement Primitives (\textbf{DMP}) model~\cite{schaal2006dynamic} is a well-known LfD approach useful for imitating a single demonstrated trajectory. 
Recent works on deep DMP (\textbf{d-DMP}) show a Deep NN learning to generate the parameters of a DMP model from an image~\cite{ridge2020training, pervez2017learning}. 
For instance, \cite{ridge2020training} propose a neural network that is trained to output the parameters of DMP \cite{schaal2006dynamic}. \cite{pervez2017learning} also use deep NNs to learn the forcing terms of the DMP model for visual servoing. 
A distribution of a set of demonstrations expresses variability of the task executions which is captured by Prbabilistic Movement Primitives (ProMP)~\cite{paraschos2018using,paraschos2013probabilistic}.
However, it fails to capture the relation between visual information and trajectory variations, hence, it misses the generalisation capability~\cite{rueckert2015extracting}. 

Breast palpation movements is hard to program and is needed for ARBP~\cite{scimeca2020structuring, herzig2018variable}. Our proposed method can address the existing challenges in motion planning for Breast Palpation, such as (1) variations across different palpation executions; (2) variability across different palpation paths; (3) relation between the visual information and demonstrated trajectories. 

In contrast to d-DMP~\cite{pervez2017learning, ridge2020training} being deterministic -- i.e. they cannot capture the distribution of trajectories and are designed only to generalise to initial and goal points -- and ProMP, which only generalise to initial, goal and via points, our approach successfully maps images of breast phantom in robot workspace to robot joint space trajectories. 
Deep movement primitives utilises deep Neural Networks (NNs) for feature extraction and subsequent generation of the weights defining a ProMP trajectory. 
Our customised loss function is defined as the distance between the Ground Truth (GT) and the trajectory generated by the predicted weights, rather than being the distance between GT and the predicted weights themselves. We implemented several different network and LfD architectures. The results obtained by our framework show its effectiveness for breast palpation tasks in a practical and real-world use case. 
We consider two sub-tasks for breast palpation. \emph{Reach to Palpate} (\textbf{RTP}): RGB-D images of the breast phantom are used to generate trajectory weights that control the robot joints to reach the starting point of palpation on the breast phantom. \emph{Wedges Palpation Path} (\textbf{WPP}): deep-MP is used to generate WPP shown in Fig.~\ref{fig:wedgepattern}) from a dataset of demonstrations. 
This framework will be integrated into a motion control in future works. 

\section{Problem Formulation}

Consider a set of $N_{\mathrm{tr}}$ demonstrations, which is defined as $\mathcal{T} := \{\{\mathbf{Q}^1, \mathrm{I}^1\}, \ldots, \{\mathbf{Q}^{N_{\mathrm{tr}}}, \mathrm{I}^{N_{\mathrm{tr}}}\}\}$ where $\mathbf{Q}^n$, $n=1,\ldots,N_{\mathrm{tr}}$, are the joint space trajectories and $\mathrm{I}^{n}$ is RGB-D images taken from the corresponding robot's workspace. For a single joint, we define a trajectory as the ordered set $\mathbf{q} := \left\{q_t\right\}_{t=1,\ldots,T}$, where $q_t\in\mathbb{R}$ is the joint position at sample $t$, and $\mathbf{Q} := \{\mathbf{q}_1, . . ., \mathbf{q}_{N_{\mathrm{joint}}}\}$ where $N_{\mathrm{joint}}$ is the number of the joints of the manipulator.

\begin{figure*}[tb] 
\centering
 \makebox[\textwidth]{\includegraphics[width=\linewidth]{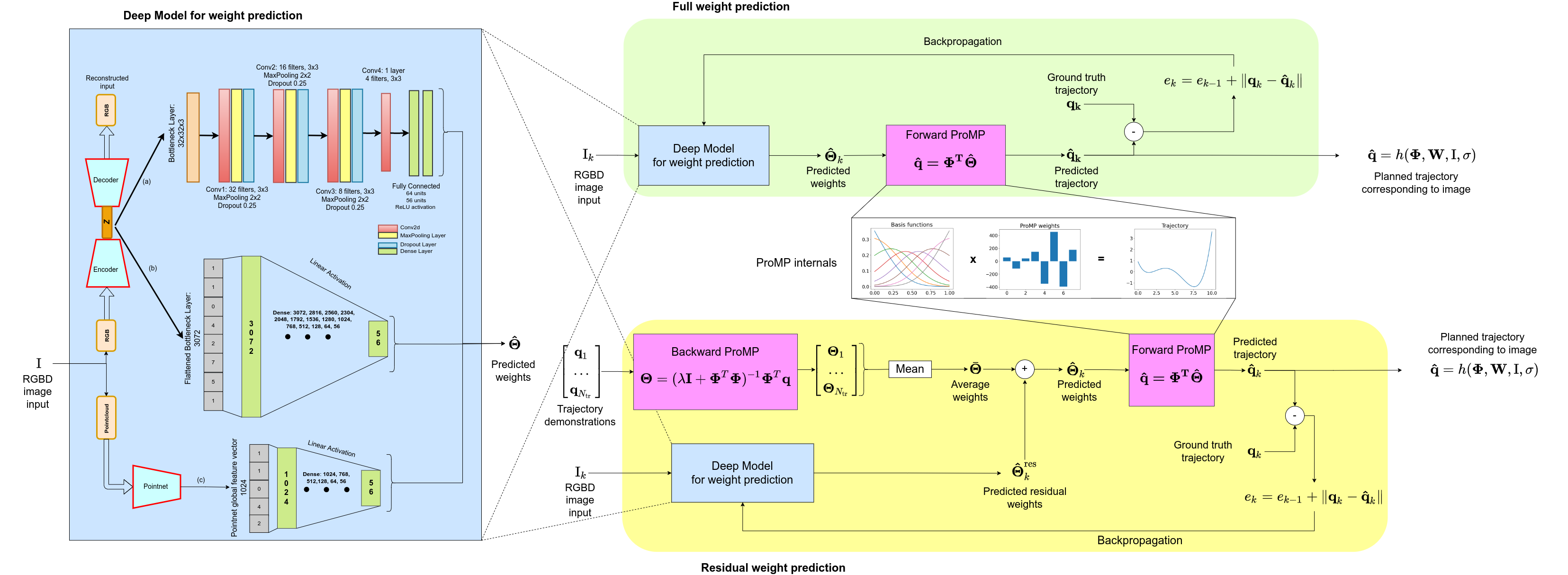}}
  \caption{{\small The blue box contains three NN models learning the weight of ProMP. RGB data is passed through an autoencoder, producing a bottleneck representation of the input $Z$, which is then fed into a CNN (a) or flattened and fed into a Fully Connected network (b). Alternatively, the depth data from the image, in the form of vector of $(x,y,z)$ coordinates of the pointcloud, is fed to a PointNet network (c) and the resulting feature vector is passed through a dense network.
  All models are used to predict the full ProMP weights (green box - Alg. \ref{algo1}) or just the residual with respect to the mean (yellow box - Alg. \ref{algo2}).
  Fig. A.4 in the Appendix is an higher resolution version of this image.}}
  \label{fig:schem_d-ProMP}
\end{figure*}

\paragraph{Probabilistic Movement Primitives (ProMP)} In order to define a distribution over trajectories, We first model a trajectory with an observation uncertainty added to the following deterministic model~\cite{paraschos2013probabilistic}:
\begin{equation}\label{eq:joint}
q = \sum_{i=1}^{N_{\mathrm{bas}}}\theta_i\psi_i(z(t)) + \epsilon_q
\end{equation} 

\noindent where $\psi_i$ are basis functions (usually Gaussian~\cite{bishop2006pattern}) evaluated at $z(t)$. $z$ is a phase function that allows time modulation. If no modulation is required, then $z(t) = t/f$, where $f$ is the sampling frequency. $\theta_i\in\mathbb{R}$ are weights, and $\epsilon_q$ adds zero-mean Gaussian observation noise with variance $\Sigma_q$. 

For stroke-like movements, the following normalised Gaussian basis functions are used:
\begin{equation}
    \psi_i(t):=\frac{b_i(z(t))}{\sum_{j=1}^{N_{\mathrm{bas}}}b_j(z(t))}
    \label{eq:norm1}
\end{equation}
\noindent where 
\begin{equation}
    b_i(z(t)):=\mathrm{exp}\left(-\frac{(z(t)-c_i)^2}{2h}\right)
    \label{eq:norm2}
\end{equation}
We can also write eq. \eqref{eq:joint} in a matrix form, as follows:
\begin{equation}
    \mathrm{q}_t = \mathbf{\Psi}_t^{T}\mathbf{\Theta} + \epsilon_q
    \label{eq:matpromp}
\end{equation}
\noindent where
$\mathbf{\Psi}_t:=(\psi_1(z(t),\ldots,\psi_{N_{\mathrm{bas}}}(z(t))\in\mathbb{R}^{{N_{\mathrm{bas}}}\times 1}$, $\mathbf{\Theta}:=(\theta_1,\ldots,\theta_{N_{\mathrm{bas}}})\in\mathbb{R}^{{N_{\mathrm{bas}}}\times 1}$, and we also define $\mathbf{\Omega}:=(\mathbf{\Theta}_1,\ldots, \mathbf{\Theta}_{N_{\mathrm{joint}}})\in\mathbb{R}^{{N_{\mathrm{bas}}} N_{\mathrm{joint}} \times 1}$ and $\mathbf{\Phi}:=\left[\mathbf{\Psi}_1,\ldots,\mathbf{\Psi}_T\right]^T\in\mathbb{R}^{T\times N_{\mathrm{bas}}}$.

From eq. \eqref{eq:joint}, it follows that the probability of observing $q_t$ is given by:
\begin{equation}
    p(q_t|\mathbf{\Theta})=\mathcal{N}\left(q_t\,\big|\,\mathbf{\Psi}^{T}_t\mathbf{\Theta}, \mathbf{\Sigma}_q\right)
    \label{eq:joint_time}
\end{equation}

Since $\Sigma_q$ is the same for every time step, the values $q_t$ are taken from independent and identical distributions, i.i.d. Hence, the probability of observing a trajectory $\mathbf{q}$ is given by:
\begin{equation}
    p(\mathbf{q}|\mathbf{\Theta}):=\prod_{t=1}^Tp(q_t|\mathbf{\Theta})
    \label{eq:joint_}
\end{equation}

However, since parameters $\mathbf{\Theta}$ are to be learnt from data, we also assume such parameters are taken from a distribution $\mathbf{\Theta}\sim p(\mathbf{\Theta}|\rho)=\mathcal{N}(\mathbf{\Theta}|\mu_{\mathbf{\Theta}},\mathbf{\Sigma}_{\mathbf{\Theta}})$. We therefore would like to have a predictive distribution of $q_t$ which does not depend on $\mathbf{\Theta}$, but on $\rho:=(\mu_{\mathbf{\Theta}},\mathbf{\Sigma}_{\mathbf{\Theta}})$. This is done by marginalising $\mathbf{\Theta}$ out in the distribution as follows:
\begin{eqnarray}\label{eq:marginal}
p(q_t|\rho)&=&\int\mathcal{N}\left(q_t\,\big|\,\mathbf{\Psi}^{T}_t\mathbf{\Theta},\,\Sigma_q\right)\mathcal{N}\left(\mathbf{\Theta}\,\big|\,\mu_{\mathbf{\Theta}},\,\Sigma_{\mathbf{\Theta}}\right)\mathrm{d}\mathbf{\Theta}
\nonumber\\
&=&\mathcal{N}\left(q_t\,\big|\,\mathbf{\Psi}^{T}_t\mathbf{\Theta},\,\Sigma_q+\mathbf{\Psi}^{T}_t\mathbf{\Sigma}_{\mathbf{\Theta}}\mathbf{\Psi}_t\right)
\end{eqnarray}

\begin{algorithm}[b!]
\caption{Deep MP}
\label{algo1}
\begin{algorithmic}[1]
\item[\textbf{Input}: NN architecture $h$, ProMP basis functions $\mathbf{\Phi}$,]
\item[image $\mathrm{I}$, training set trajectories $\mathbf{q}$, activation function $\sigma$]
\item[\textbf{Output}: NN weights $\mathbf{W}$, predicted trajectory $\mathbf{\hat{q}}$ ]
\item[{\small \textbf{Note}: This pseudo code is for single joint trajectory $\mathbf{\hat{q}}$.}] \item[{\small Generalising it for all joints $\mathbf{\hat{Q}}$ is straightforward.}]
\item[---------------------------------]

\STATE $\mathit{Dataset}$: $\mathcal{T} \gets \{\mathit{\mathbf{q}, \mathrm{I}}\}_{1, . . .,N_{\mathrm{tr}}}$
\STATE $\mathit{Initialise DeepModel:}\: \mathbf{\hat{\Theta}} \gets  h  (\mathbf{W}, \mathbf{\mathrm{I}}, \mathbf{\sigma}) $ eq.~\eqref{eq:dpromp} \newline \nonumber as per Fig.~\ref{fig:rtpresult}: either CNN (2-D) or FCN (1-D);
\STATE $\mathit{Initialise ProMP}$: $\: \mathbf{\hat{q}} \gets \mathbf{\Phi}^T \mathbf{\hat{\Theta}}$  (eq.~\eqref{eq:matpromp})
\STATE $\mathit{RMSE} \gets \: e = \| \mathbf{\hat{q}} - \mathbf{q} \|$

\WHILE {$(\mathit{e} > \epsilon$)}
    \FORALL{$\{\mathit{\mathbf{q}, \mathrm{I}}\}\in  \mathcal{T}$}
        \STATE $\textsf{\textsc{ForwardPropagation:}}  \: \mathbf{\hat{\Theta}}_k =  h (\mathbf{W}_k, \mathbf{\mathrm{I}}_k, \mathbf{\sigma}) $
        \STATE $\textsf{\textsc{ForwardProMP:}}  \:  \mathbf{\hat{q}}_k = \mathbf{\Phi}^T \mathbf{\hat{\Theta}}_k $  (eq.~\eqref{eq:matpromp})
        \STATE $\textsf{\textsc{RmseJointLoss:}}  \:  \mathit{e}_k = \mathit{e}_{k-1} + \| \mathbf{\hat{q}}_k - \mathbf{q}_k \|$
    \ENDFOR
    \STATE $\textsf{\textsc{BackPropagation:}}$  $\mathbf{W}_{k+1} \gets  \{\mathbf{W}_k, \frac{\partial{\mathit{e}_k}}{\partial{\mathbf{W}_k}}\}$
\ENDWHILE

\STATE deep-MP: $\hat{\mathbf{q}} = h(\mathbf{\Phi}, \mathbf{W}, \mathbf{\mathrm{I}}, \mathbf{\sigma})$

\STATE \textbf{end}
\end{algorithmic}
\end{algorithm}

\paragraph{deep-MP weights learning:}
The weights of ProMP models are conventionally learned from demonstrations where they can be later adapted according to different trajectory reproduction needs, e.g. (1) the initial/goal point of the desired trajectory are set, or (2) some via points are determined based on the problem constrains.
Computing the via/start/goal points based on visual sensory information of robot's workspace needs hand-designed and task- or robot's workspace-specific features. This is effective and handy in some robotic tasks like simple pick-and-place, but it is too complex for breast palpation, e.g., in which the task trajectory and the geometry of the breast are related. Two following deep-MP models learn the relation between visual sensory information and joint trajectories.

\paragraph{(deep-MP):} Instead of learning the weights of the ProMP, a deep model-- that can be a CNN, FC or PointNet model-- captures the correlation between the visual sensory information and ProMP weights as per eq. \eqref{eq:dpromp}. The algorithm is described in Alg.~\ref{algo1}. Moreover, a schematic of the algorithm is shown in Fig.~\ref{fig:schem_d-ProMP} in the green block at top right).

For a neural network similar to Fig. \ref{fig:schem_d-ProMP}, network's behaviour can be described by the following:
\begin{equation}
\mathbf{\Theta}_k =  h_k (\mathbf{W_k}, \mathbf{\mathrm{I}_k}, \mathbf{\sigma_{k}}) + v_{k} \label{eq:dpromp}
\end{equation}

Equation \eqref{eq:dpromp}, known as observation equation, shows that the network’s target vector $\mathbf{\Theta}_k$ is equivalent to a nonlinear function $h_k$ of the input image $\mathrm{I}_k$, the weight parameter $\mathbf{W}_k$, the node activation $\sigma_k$ and the observation/measurement
noise $v_k$. We consider the measurement noise to be a zero-mean white noise with covariance given by $E[v_k v_l^T] = R_k$.

\begin{figure}[tb!]
    \begin{subfigure}{1\columnwidth}
        \centering
        \includegraphics[width=.99\linewidth]{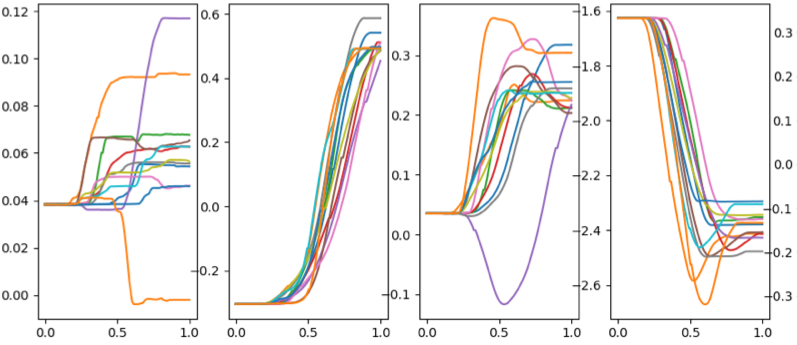}
        \caption{{\small Joint Space trajectory}}
        \label{fig:schNI2}
    \end{subfigure}%
    \\
    \begin{subfigure}{1\columnwidth}
        \centering
        \includegraphics[width=.99\linewidth]{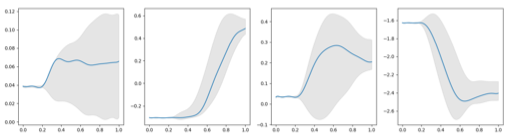}
        \caption{{\small Reproduction mean trajectory and corresponding variance}}
        \label{fig:insertion}
    \end{subfigure}%
    \caption{(a) Samples of RTP-RGBD demonstrated trajectories from left/joint 1 to right/joint 4 -- trajectories of joint 5, 6 and 7 are not shown here; (b) the computed mean (solid blue lines)-- used for the residual deep-MP implementation-- and representative variations of the distribution (shaded grey areas).}
    \label{fig:palpation_set}
\end{figure}

\begin{algorithm}[b!]
\caption{Residual deep MP}
\label{algo2}
\begin{algorithmic}[1]
\item[\textbf{Input}: NN architecture $h$, ProMP basis functions $\mathbf{\Phi}$,]
\item[image $\mathrm{I}$, training set trajectories $\mathbf{q}$, activation function $\sigma$] 
\item[\textbf{Output}: NN weights $\mathbf{W}$, predicted trajectory $\mathbf{\hat{q}}$]
\item[{\small \textbf{Note}: This pseudo code is for single joint trajectory $\mathbf{\hat{q}}$.}] \item[{\small Generalising it for all joints $\mathbf{\hat{Q}}$ is straightforward.}]
\item[--------------------------------]
\STATE $\mathit{Dataset}$: $\mathcal{T} \gets \{\mathit{\mathbf{q}, \mathrm{I}}\}_{1, . . .,N_{\mathrm{tr}}}$
\STATE $\mathit{InitWeights}$: $\{\mathbf{\Theta}_1, ..., \mathbf{\Theta}_{N_ {\mathrm{tr}}}\} \gets \mathbf{\Phi} \cdot \{\mathbf{q}_1, ..., \mathbf{q}_{N_ {\mathrm{tr}}}\}$
\STATE $\mathit{InitAverages}: \bar{\mathbf{\Theta}} $ $\gets$ $ \mathrm{mean}(\mathbf{\Theta}_1, . . ., \mathbf{\Theta}_{N_{\mathrm{tr}}})$ (eq.~\eqref{eq:weight})
\STATE $\mathit{InitDeepModel:}\: \mathbf{\hat{\Theta}}^{\mathrm{res}} \gets  h (\mathbf{W}, \mathbf{\mathrm{I}}, \mathbf{\sigma}) $ eq.~\eqref{eq:dpromp} \newline \nonumber as per Fig.~\ref{fig:rtpresult}: either CNN (2-D) or FCN (1-D);
\STATE $\mathit{InitFullWeights}: \mathbf{\hat{\Theta}} = \mathbf{\hat{\Theta}}^{\mathrm{res}} + \mathbf{\bar{\Theta}}$ (eq. \eqref{eq:residual})
\STATE $\mathit{InitProMP}: \hat{\mathbf{q}} \gets \mathbf{\Phi} \mathbf{\hat{\Theta}}$ (eq.~\eqref{eq:matpromp})
\STATE $\mathit{RMSE} \gets \: e = \|{\mathbf{\hat{q}}} - \mathbf{q}\|$

\WHILE {$(\mathit{e} > \epsilon$)}
    \FORALL{$\{\mathit{\mathbf{q}, \mathrm{I}}\}\in  \mathcal{T}$}
        \STATE $\textsf{\textsc{ForwardPropagation:}}  \: \mathbf{\hat{\Theta}}_k^{\mathrm{res}} =  h (\mathbf{W}_k, \mathbf{\mathrm{I}}_k, \mathbf{\sigma}) $
        \STATE $\textsf{\textsc{FullWeights:}} \:  \mathbf{\hat{\Theta}} = \mathbf{\hat{\Theta}}^{\mathrm{res}} + \mathbf{\bar{\Theta}}$ (eq. \eqref{eq:residual})
        \STATE $\textsf{\textsc{ForwardProMP:}}  \:  \mathbf{\hat{q}}_k = \mathbf{\Phi}^T \mathbf{\hat{\Theta}}_k $  (eq.~\eqref{eq:matpromp})
        \STATE $\textsf{\textsc{RmseJointLoss:}}  \:  \mathit{e}_k = \mathit{e}_{k-1} + \|\mathbf{\hat{q}}_k - \mathbf{q}_k \|$
    \ENDFOR
    \STATE $\textsf{\textsc{BackPropagation:}} \mathbf{W}_{k+1} \gets  \{\mathbf{W}_k, \frac{\partial{\mathit{e}_k}}{\partial{\mathbf{W}_k}}\}$
\ENDWHILE

\STATE deep-MP residual: $\hat{\mathbf{q}} = h(\mathbf{\Phi}, \mathbf{W}, \mathbf{\mathrm{I}}, \mathbf{\sigma})$

\STATE \textbf{end}
\end{algorithmic}
\end{algorithm}

\noindent where $h_k$ is a nonlinear deep model mapping the image $\mathrm{I}_k$ taken by robot's camera at a home position (see Fig.~\ref{fig:setup0}) to the ProMP weights. The weights then generate the corresponding trajectories using eq.~\eqref{eq:matpromp}. 
\paragraph{(Residual deep-MP):} usually a set of demonstrated trajectories convey information about the presented behaviour regardless of the scene. GMM, GP and ProMP have been used to encode such information in a probabilistic model that can be expressed as a mean and distribution (see Fig.~\ref{fig:palpation_set}). In order to improve the performance of the deep-MP, we propose a deep model which learns the correlation between the input image and residual trajectories, i.e. difference between the mean and demonstrated trajectories. Hence, the deep model is required to learn a much simpler mapping since part of the complexities are captured by the mean trajectory (see the yellow block at bottom right in Fig.~\ref{fig:schem_d-ProMP}). 
First, we learn  to fit a ProMP trajectory $\hat{\mathbf{Q}}$ to the demonstrated trajectories using least squares optimisation. 

Hence, we can compute the mean weights $\bar{\mathbf{\Theta}}$ and mean joint space trajectories $\mathbf{q}$ (see Fig.~\ref{fig:palpation_set}) by maximising the likelihood 
as per eq.~\eqref{eq:weight}. 

\begin{equation} 
\begin{split}
{\mathbf{\Theta}}^{n} = (\lambda \mathbf{I} + \mathbf{\Phi}^T \mathbf{\Phi})^{-1} \mathbf{\Phi}^T \mathbf{q}^{n},\,\forall n=1,\ldots,N_{\mathrm{tr}}\\
\bar{\mathbf{\Theta}} = \mathbb{E}([\mathbf{\Theta}^1, . . .,\mathbf{\Theta}^{N_{\mathrm{tr}}}])
\end{split}
\label{eq:weight}
\end{equation}
\noindent where $\mathbf{q}^n:=\left(q_1,\ldots,q_T\right)\in\mathbb{R}^{T\times 1}$ is the vectorized form of single-joint values in trajectory $n$, and $\lambda$ is a regularising term used to avoid over-fitting in the original optimisation objective. 
Then the deep-MP model learns the correlation between the residuals of the trajectories and the visual sensory information. 

\begin{equation} \label{eq:residual}
 \mathbf{\Theta}^{n} = \mathbf{\Theta}^{\mathrm{res},n} +\bar{\mathbf{\Theta}}
\end{equation}
\noindent where $\mathbf{\Theta}^{n}$ and $\mathbf{\Theta}^{\mathrm{res},n}$ are, respectively, the full and residual weights of the ProMP model for $N_{\mathrm{tr}}$ demonstrated trajectories.  
We can use the deep-MP model presented in eq.~\eqref{eq:dpromp} to learn $\mathbf{\Theta}^{\mathrm{res},n}$ which will be added to the $\bar{\mathbf{\Theta}}$ to form the ProMP corresponding with $\{{\mathbf{Q}}^{n}, {\mathrm{I}}^{n}\}$, as per eq.~\eqref{eq:matpromp}.

\section{Hardware setup and data collection}

Our experimental setup consists of a 7-DoF Panda robotic arm manufactured by Franka Emika.
An Intel RealSense D435i RGB-D camera is mounted on the wrist of the arm. We also use a tactile finger consisting of a 6x4 uSkin Xela magnetic-based tactile sensor for RTP-RGBD and WPP data collection. This sensor is firmly connected to the left finger link of the gripper using a 3D printed mount. Although the reading of the tactile sensing is not used in this study, we will use it for future study of palpation motion control.

We have obtained three data sets: (1) we collected reach-to-palpate (RTP) dataset, called RTP-RGB, in a mock study; (2) Reach-to-palpate dataset 2 called RTP-RGBD, and (3) Wedged-palpation-path dataset called WPP.

\paragraph{Reach-to-palpate (RTP)}  

\begin{figure}[tb!]
    \centering
    \includegraphics[width=1.0\linewidth]{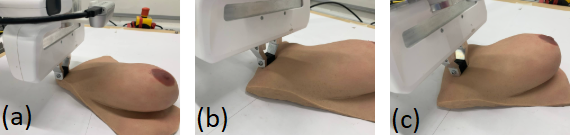}
    \caption{{\small RTP-RGB Datasets: CNN--deep-MP Model tested on unseen Breast Phantom configurations. (a) shows precise task execution whereas (b) and (c) have a larger errors as they belong to regions with different sample densities (a, b and c belong to region 1, 2 and 4, respectively in Fig. A.3 of Appendix). The distance between the Robot EE and the corner of the breast phantom--the desired touching point demonstrated-- is because of difference in sample density of the corresponding regions.}}
    \label{fig:rtpresult}
\end{figure}

(i) RTP-RGB: Our mock study includes the RTP-RGB data collection. For each sample in this dataset, the robotic arm starts from a fixed home pose as shown in Fig. \ref{fig:setup0}. At this home pose, the camera takes an RGB image of the breast phantom; and then the robot is manually moved to the corner of the breast phantom in kinesthetic teaching mode as shown in Fig. \ref{fig:rtpresult}. This dataset contains 500 samples, i.e. the robot at home configuration takes RGB images of breast phantom at a random position on the table. We have trained the CNN and FC deep-MP model, where they yield 0.0108 and 0.0118 [Radian$^2$] errors in joint space and 39.7 mm and 46.8 mm errors in task space. 

Full details of the RTP-RGB dataset and the obtained results are described in Appendix. The results obtained by this dataset suggest (1) we need a more structured dataset to better understand the impact of samples density on the results; (2) the depth data may be relevant; (3) a more challenging task is needed to showcase the effectiveness of the approach.   

\begin{figure}[tb!]
    \centering
    \includegraphics[width=.9\linewidth]{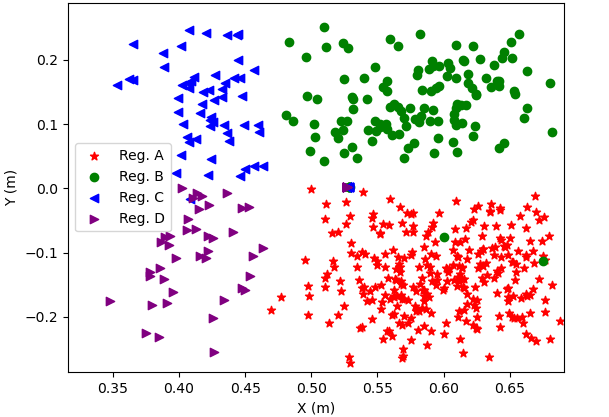}
    \caption{{\small XY coordinates of the end-effector when the robot reaches the start point of the palpation in RTP-RGBD dataset. }}
    \label{fig:xy_ee_final_rtp2}
\end{figure}

(ii) RTP-RGBD: the setup for the following data collection is the same as the one in RTP-RGB dataset. Nonetheless, we collected RGB and depth data for each sample and the robot is moved by joint space motion planning to the nipple of the breast phantom. We consider 4 regions for data collection as shown in Fig.~\ref{fig:xy_ee_final_rtp2}: Region A, B, C and D. After each sample collection, the breast phantom was moved to a new location within the region boundary to create uniform distribution for each region with different densities as shown in Fig. \ref{fig:xy_ee_final_rtp2}. A total of 545 samples were collected with 292, 128, 73 and 52 samples in region A, B, C, and D respectively. 

\begin{figure}[tb!]
\centering
    \begin{subfigure}[b]{0.1\textwidth}
        \centering
        \includegraphics[width=\textwidth]{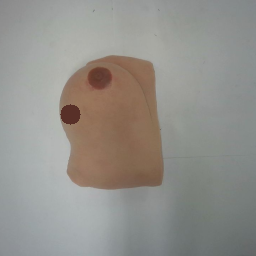}
        {{\small Config. I}}    
    \end{subfigure}
    \begin{subfigure}[b]{0.1\textwidth}  
        \centering 
        \includegraphics[width=\textwidth]{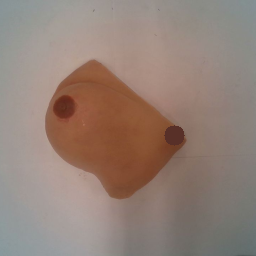}
        {{\small Config. II}}    
    \end{subfigure}
    \begin{subfigure}[b]{0.1\textwidth}   
        \centering 
        \includegraphics[width=\textwidth]{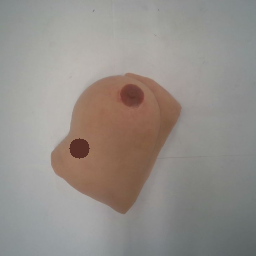}
        {{\small Config. III}}    
    \end{subfigure}
    \begin{subfigure}[b]{0.1\textwidth}   
        \centering 
        \includegraphics[width=\textwidth]{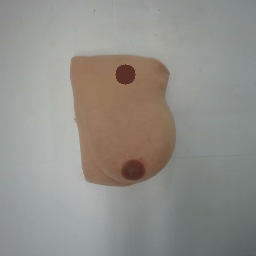}
        {{\small Config. IV}}    
    \end{subfigure}
    \caption{Samples of WPP data set: the brown disk on images show the desired end-points of palpation path 5, 1, 4 and 3 at configuration I, II, III and IV, respectively. The starting point for each demonstrated palpation is the nipple.}
    \label{fig:wpp_configs}
\end{figure}

\paragraph{Wedges Palpation Path (WPP)} 
After moving the robot tactile finger to the nipple of the breast phantom, the robot needs to follow the palpation path. The robot is moved using kinesthetic-teaching-mode from nipple along to the edge of the phantom similar to WPP shown in Fig.~\ref{fig:wedgepattern}. 
Synchronised robot full state, tactile sensor readings, and joint trajectory are recorded. 
31 palpation trials for every 7 WPPs (Fig.~\ref{fig:wedgepattern}) and four different phantom configurations  (Fig. \ref{fig:wpp_configs}) are recorded. A total of 868 palpation samples were collected in WPP dataset.

\section{Experiments and Results}
To validate our hypothesis, i.e. deep-MP can produce accurate trajectory/path for breast palpation, we performed different experiments using the collected dataset RTP-RGBD and WPP in this proof-of-concept study. Although we only have one breast model and we cannot claim the generalisation across different breast geometry, the palpation paths varies across different palpation experiments in WPP experiments due to variations in the palpation terminal points. 

Autoencoder as feature extractor: Using the RGB images, we trained an autoencoder on the 500 images for feature extraction, by using the bottleneck layer of the trained autoencoder. We then trained two models that use the learned features from the bottleneck layer to predict the joint weights that takes the robot to the palpation starting point (see Fig.~\ref{fig:schem_d-ProMP}).

Convolutional Neural Network (CNN) for mapping from features to ProMP weights Model: The bottleneck layer was an image of 32x32x3 which was passed through a CNN architecture to predict the joint weights.
We implemented two versions of this model, one using Alg. \ref{algo1} that predicts the full ProMP weights and one using Alg. \ref{algo2} that predicts the residual ProMP weights with respect to the mean trajectory for any given region.

\begin{table*}[tb!]
    \centering
    \begin{threeparttable}[b]
    \begin{tabular}{@{}c@{\hskip 0.8cm}cccc@{\hskip 0.8cm}cccc@{}}
        \toprule
        \multirow{3}{*}{Model} & \multicolumn{4}{c}{MSE$^*$}& \multicolumn{4}{c}{Absolute Error$^{**}$} \\
           & Reg. A (292 samples) & B (128) & C (73) & D (52) & A (292) & B (128) & C (73) & D (52)\\
        \midrule
        CNN residual deep-MP & \textbf{0.0009} & \textbf{0.0030} & 0.0051 & 0.0047 & \textbf{24.6} & \textbf{41.7} & 72.1 & 72.5 \\
        CNN deep-MP & 0.0016 & 0.0031 & \textbf{0.0037} & \textbf{0.0041} & 41.1 & 48.0 & \textbf{55.2} & \textbf{58.5} \\
        FC deep-MP & 0.0047 & 0.0051 & 0.0058 & 0.0082 & 47.2 & 63.3 & 69.5 & 83.8 \\
        PointNet deep-MP & 0.0095 & 0.0169 & 0.0233 & 0.0191 & 75.8 & 192.3 & 226.8 & 167.7 \\
        d-DMP & 0.0117 & 0.0109 & 0.0172 & 0.0226 & 84.8 & 66.5 & 113.8 & 132.6 \\
        \bottomrule
    \end{tabular}
    \begin{tablenotes}
        \item \small $^{*}$ MSE in rad${}^2$; $^{**}$ Absolute error in mm;
    \end{tablenotes}
    \end{threeparttable}
    \caption{Evaluation of RTP models with Average MSE (AveMSE) of the joint trajectory in each region of RTP-RGBD data set (columns at left); Evaluation of RTP models with Average Euclidean Distance (AveED) in cartesian coordinates of the end effector position at the end of the trajectory in each region of RTP-RGBD data set (columns at right).}
    \label{table:rtprgbd_mse_ed}
\end{table*}

\begin{table*}[tb!]
    \centering
    \begin{threeparttable}[b]
    \begin{tabular}{@{}c@{\hskip 0.4cm}cccc@{\hskip 0.8cm}cccc@{}}
        \toprule
        \multirow{3}{*}{Model} & \multicolumn{4}{c}{MSE$^*$}& \multicolumn{4}{c}{Absolute Error$^{**}$} \\
           & Config.I & Config.II & Config.III & Config.IV & Config.I & Config.II & Config.III & Config.IV\\
        \midrule
        CNN deep-MP & 0.0373 & \textbf{0.0245} & \textbf{0.0072} & \textbf{0.0420} & 67.4 & \textbf{57.6} &\textbf{53.9}& 63.1\\
        CNN residual deep-MP & 0.0380 & 0.0763& 0.0662& 0.0465 & 61.3 & 58.2 & 89.3 & 83.6\\
        d-DMP & 1.0541 & 1.4307 & 1.7214 & 1.7013 & 487.0 & 499.9 & 399.8 & 407.9\\
        \midrule
        WPP4 CNN deep-MP &  \textbf{0.0340}& 0.0617& 0.0441& 0.0756&  \textbf{63.8}&  112.9& 113.5& \textbf{61.4}\\
        \bottomrule
    \end{tabular}
    \begin{tablenotes}
        \item \small $^{*}$ MSE in rad${}^2$; $^{**}$ Absolute error in mm. 
    \end{tablenotes}
    \end{threeparttable}
    \caption{Evaluation of CNN deep-MP and CNN residual deep-MP Model with AveMSE and Absolute Error at each configuration for WPP9 data set (middle row). Evaluation of CNN deep-MP for WPP4 dataset (bottom row).}
    \label{table:wpp_mse_ed}
\end{table*}

Fully Connected (FC) Model for mapping from features to ProMP weights: The bottleneck layer was flattened to a 1-D vector and then fed into stack of dense layers to predict the joint weights at the last layer.

PointNet for mapping the pointcloud coordinates to ProMP weights: In addition to the CNN and FC models described above, we also trained deep models using the pointcloud coordinates. The vector of Cartesian coordinates was fed into a pre-trained PointNet network~\cite{qi2017pointnet}, from which the global feature vector was extracted and fed into a smaller Fully Connected network to predict the joint weights (Blue block in Fig.~\ref{fig:schem_d-ProMP}).\\

For all models, the dataset is split into 85\% and 15\% for training and testing, respectively, and 25\% of training data are used for cross-validation. We use Tensorflow framework for training with Adam optimiser using a learning rate of 0.001 and a batch size of 32. Our models were trained using 150 epochs with a callback function to terminate training when over-fitting.

\paragraph{Loss function} We have implemented our custom loss function using the deep-MP model. We denote the ground truth of our joints weights as $\mathbf{\Theta}_{\mathrm{gt}}\in\mathbb{R}^{N_{\mathrm{bas}}\times 1}$ and the corresponding predicted ones as $\mathbf{\Theta}_{\mathrm{ps}}\in\mathbb{R}^{N_{\mathrm{bas}}\times 1}$. Our loss function is the root mean squared error between the trajectories generated by $\mathbf{\Theta}_{\mathrm{gt}}$ and $\mathbf{\Theta}_{\mathrm{ps}}$. The loss of a predicted vector $\mathbf{\Theta}_{\mathrm{ps}}$ is then given by:

\begin{equation}
L(\mathbf{\Theta}_{\mathrm{ps}}, \mathbf{\Theta}_{\mathrm{gt}}) := \sqrt{\left(\frac{1}{T}\right)\sum_{t=1}^{T}(q_{\mathrm{gt},t}-q_{\mathrm{ps},t})^{2}}
\end{equation}

where $q_{\mathrm{gt},t} =\mathbf{\Phi}^T_t\mathbf{\Theta}_{\mathrm{gt}}$ and $q_{\mathrm{ps},t} = \mathbf{\Phi}^T_t\mathbf{\Theta}_{\mathrm{ps}}.$ 

For our experiments, the number of samples was $T=150$.

\paragraph{Metrics} To measure the performance of the approach, $N_{\mathrm{test}}$ test samples are evaluated. Let $\mathbf{\Theta}_{\mathrm{ps}}^n\in\mathbb{R}^{N_{\mathrm{bas}}\times 1}$ be the predicted weights of test sample $n$. We consider the following two metrics.

\noindent (i) Average Mean Squared Error (AveMSE) between the trajectories generated from the predicted weights and ground truth.

\begin{equation}
    \mathrm{AveMSE} := \frac{1}{N_{\mathrm{test}}}\sum_{n=1}^{N_{\mathrm{test}}}\left[L(\mathbf{\Theta}_{\mathrm{ps}}^n, \mathbf{\Theta}_{\mathrm{gt}}^n)\right]^2
    \label{eq:avemse}
\end{equation}

where $N_{\mathrm{test}}$ is the number of samples in the test set. 

\noindent (ii) Average Euclidean Distance (AveED) between the position of the end-effector at the last configuration, $t = T$ generated by deep-MP model and the ground truth. 

\begin{equation}
    \mathrm{AveED} := \frac{1}{N_{\mathrm{test}}}\sum_{n=1}^{N_{\mathrm{test}}}\left|\mathbf{r}_{\mathrm{ee,ps}}^n-\mathbf{r}_{\mathrm{ee,gt}}^n\right|
    \label{eq:aved}
\end{equation}

\noindent where, $\mathbf{r}_{\mathrm{ee,gt}}^n,\,\mathbf{r}_{\mathrm{ee,ps}}^n\in\mathbb{R}^{3\times 1}$ are, respectively, the ground truth and predicted position vectors of the end-effector for sample $n$, at $t=T$.

The CNN network architectures and the loss function for the WPP and RTP are the same,  except the last units of the dense layer in CNN-RTP is replaced with $70$ units in CNN-WPP. Moreover, the images input to autoencoder are marked with the target point as shown in Fig. \ref{fig:wpp_configs}. 
The training parameters for WPP and RTP experiments are the same. However, 200 epochs was used in WPP. 

\paragraph{d-DMP} \cite{ridge2020training} recently proposed deep DMP. We also implemented d-DMP both for RTP and WPP tasks with our best CNN (see Appendix for implementation details).
The results obtained by d-DMP and deep-MP are shown in  Table~\ref{table:wpp_mse_ed} for WPP experiments illustrating our deep-MP models performance is much better than the d-DMP ones in all the cases.

While the CNN deep-MP outperforms the FC, d-DMP and PointNet deep-MP model in RTP-RGBD dataset in RTP experiments (Table~\ref{table:rtprgbd_mse_ed}), the performance of d-DMP is better than just some cases of PointNet deep-MP. CNN can capture more complex mapping than FC model and PointNet deep-MP; hence, it is by far better than others (the trajectories generated by  the CNN models are shown in Fig. A.5 and A.6 of Appendix).

We considered different configurations of training and test set for WPP task conducting a series of experiments -- $\{$WPP1, $\dots$, WPP10$\}$ (see Appendix for details).
Two experiments with interesting results are presented in Table~\ref{table:wpp_mse_ed}. 
Our results (see Table A.4 in Appendix) show the best model performance is obtained when including all the palpation path, as it was done in WPP9 experiment. The results of WPP9 are also presented in the middle row of Table~\ref{table:wpp_mse_ed} showing all deep-MP models outperforming d-DMP by a large margin. 
Another interesting observation is presented in the bottom row of Table~\ref{table:wpp_mse_ed} for WPP4 where -- similarly to WPP1-3 -- only palpation 1,4,5 are seen in training and 2,3 used for testing. The results indicate that our model is able to generalise to unseen geometry of breast palpation. 
Other eight ablation studies also suggest the effectiveness of our deep-MP model for complex palpation tasks. 

The results in Table \ref{table:rtprgbd_mse_ed} and  \ref{table:wpp_mse_ed} show that residual version of deep-MP outperforms the standard one in RTP but not quite well in WPP. We learned that RTP trajectories are more uniform than WPP ones. Moreover, the RTP has travelling distance longer than WPP with much more non-linearity. Hence, the mean trajectory across different RTP demonstrations is more informative than the WPP. i.e. the mean trajectory captures large non-linearity across the RTP trajectories, hence the CNN model requires to learn much simpler mapping. 

In spite of WPP results (in Table~\ref{table:wpp_mse_ed}), the MSE and Absolute Error of d-DMP and the variations of deep-MP reported in RTP cases (in Table~\ref{table:rtprgbd_mse_ed}) are very close. Hence, the combined results in Table~\ref{table:rtprgbd_mse_ed} and Table~\ref{table:wpp_mse_ed} suggest that d-DMP can perform relatively well in simple tasks such as RTP-- where, in contrast to WPP, RTP tasks depend only on a target point, i.e. the nipple-- whereas it yields a very poor performance in complex tasks such as WPP. 
The video attachment shows deep-MP successfully outputting the joints weights for palpation paths from the nipple to a terminal point that is set manually -- demonstrating the generalisation of our model -- or autonomously.

\section{Conclusion}

Autonomous robots for breast palpation can have significant impact on the societies health sector. One of the challenges for such technologies is robot programming which forms a very interesting scientific problem. We proposed a novel learning from demonstration method, called deep movement primitives, that maps visual information to palpation trajectories and is useful for programming a robot for autonomous palpation. 
We show the effectiveness of our approach in a series of real-robot breast phantom palpation experiments. While state-of-the-art approach fails to learn palpation due to the variations in the start, terminal points and the intermediate points, our results shows deep-MP is a suitable method for learning complex tasks, such as palpation. Our future works include use of the sensor data and creating demonstrations across different breast models that helps deep-MP generalise even better.

\section*{Ethical statement}
Cancer Research UK has funded a project in 2020 to shape new technology for breast health called \emph{ARTEMIS}~\footnote{https://bit.ly/3GFIsC3}. This proof of concept project aims to develop a breast cancer examination robot in collaboration with clinicians and with Patient and Public Involvement (PPI)~\cite{artemissurvey2020}. The consortium consists of University of Lincoln, University of Bristol and Imperial College London. Our partners are developing a soft robot tool and soft sensors for safe, non-invasive and comfortable use, building Breast phantoms based on real subjects' breasts that can give clinicians a feel of real breasts during palpation, collecting data of clinicians performing palpation, with active PPI in design and development of the technology. 

The autonomous Breast Palpation Robot (ABPR) will have a chair where patients can incline/bend forward 45 degrees and lie on the top of the robot housing and breasts go into designated holes allowing the soft sensors to touch the breast and perform the palpation. A survey of 155 women in the United Kingdom was conducted, showing them schematics of different designs of ABPR. Results indicated enthusiasm for ABPR with 92\% of respondents indicating they would use ABPR. 83\% would be willing to be examined for up to 15 minutes. GP surgery is the most popular location for ABPR. Thematic analysis of free-text responses identified the following: a) Subjects perceived ABPR has the potential to address limitations in current screening services; b) ABPR facilitates increased user choice and autonomy; c) there are ethical motivations for supporting ABPR development; d) accuracy is essential; e) integration with health services is important.
The above results are included in a paper \cite{artemissurvey2020} under revision for final publication.

ABPR collects data for clinician references and potentially reduces the number of (true negative) visits to hospitals for breast cancer examination. It is no replacement for hospital examinations. We aim to develop a cheap device safe, comfortable, reliable and accessible--especially in poor communities/countries that lack hospitals or expert clinicians.

ABPR allows recording the history of the palpation data for individuals helping patients and clinicians with precise information about any changes observed during palpation to be judged by clinicians or AI whereas judgement based on human palpation can be subjective. 

\section*{Acknowledgements}
This work was partially supported by Centre for Doctoral Training, United Kingdom (CDT) in Agri-Food Robotics (AgriFoRwArdS) Grant reference: EP/S023917/1; Lincoln Agri-Robotics (LAR) funded by Research England; and by ARTEMIS project funded by Cancer Research UK C24524/A300038.

\bibliography{reference.bib}
\end{document}